\documentclass[runningheads]{llncs}

\usepackage{eccv}

\usepackage{eccvabbrv}
\usepackage{graphicx}
\usepackage{booktabs}
\usepackage[accsupp]{axessibility}

\usepackage{orcidlink}

\usepackage{amsmath,amssymb}
\usepackage{algorithm}
\usepackage{algorithmic}

\usepackage{multirow}
\usepackage{xcolor}
\usepackage{colortbl}
\usepackage{array}

\usepackage{subcaption}

\definecolor{tableheader}{RGB}{46,134,171}
\definecolor{tablerowalt}{RGB}{245,248,250}
\definecolor{bestresult}{RGB}{46,134,171}
\definecolor{forgettinggap}{RGB}{200,50,50}

\usepackage{float}
\usepackage{wrapfig}

\begin{document}

\title{Do Vision Models Truly Forget? \\
New Findings from Representation-Level Certification of Visual Unlearning in Vertical Federated Learning}

\titlerunning{Mirage}

\author{
Zhenyu Yu\inst{1,\dagger} 
\and
Yangchen Zeng\inst{2,\dagger} 
\and
Chunlei Meng\inst{3}
\and
Guangzhen Yao\inst{4}
\and
\\Shuigeng Zhou\inst{1,*}
}

\authorrunning{Yu et al.}

\institute{College of Computer Science and Artificial Intelligence, Fudan University 
\and
School of Cyber Science and Engineering, Southeast University
\and
College of Intelligent Robotics and Advanced Manufacturing, Fudan University
\and
School of Information Science and Technology, Northeast Normal University
\\
\email{zhenyuyu@fudan.edu.cn; zeng.yc@bytedance.com; clmeng23@m.fudan.edu.cn; guangzhenyao@163.com; sgzhou@fudan.edu.cn}\\
\inst{*}~Corresponding author~~~   \inst{\dagger}~Co-first author
}

\maketitle

\vspace{-20pt}
\begin{figure}
    \centering
    \includegraphics[width=1\linewidth]{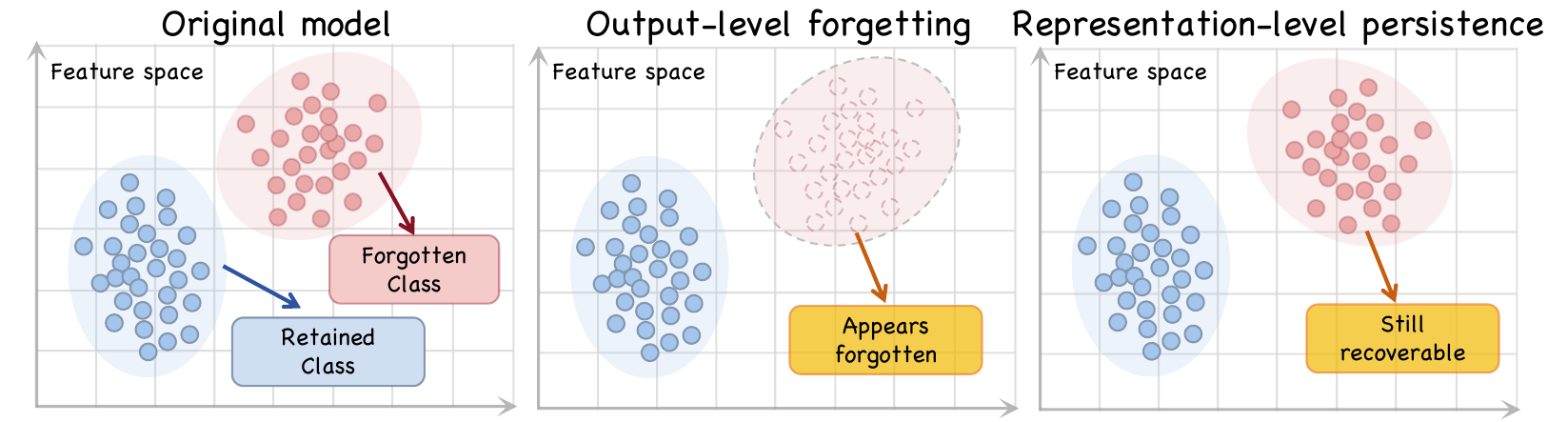}
    \vspace{-20pt}
    \caption{\textbf{Mirage reveals the illusion of forgetting.} Suppressing classifier predictions may create the appearance of successful unlearning (middle), while the underlying feature geometry remains largely unchanged. Consequently, a linear probe can still recover forgotten-label information with high accuracy (right). This mismatch between behavioral suppression and representational persistence forms the \emph{forgetting illusion}.}
    \label{fig:motivation}
\end{figure}

\vspace{-20pt}
\begin{abstract}
  Machine unlearning in Vertical Federated Learning (VFL) has attracted growing interest, yet existing methods certify forgetting solely using output-level metrics. We challenge these works by introducing \textbf{Mirage}, a representation-level auditing framework that comprises four complementary diagnostics: Linear probe recovery (LPR), centered kernel alignment (CKA), feature separability scoring, and layer-wise recovery analysis. Extensive experiments across seven datasets and seven baseline methods following recent VFL unlearning protocols reveal three key findings: \emph{(1) Forgetting gap}: methods that pass output-level certification still retain substantial class structure in their representations, with LPR exceeding the retrained baseline by up to $15.4$ points; CKA shows that these models remain structurally closer to the original than to the retrained reference, while separability scores indicate persistent geometric discrimination. \emph{(2) Unlearning trilemma}: no existing method simultaneously achieves high utility, output-level forgetting, and representation-level forgetting. \emph{(3) Class-sample asymmetry}: class-level forgetting leaves strong representational traces (LPR exceeding 96\% on several datasets), whereas sample-level forgetting is indistinguishable from chance (LPR $\approx50\%$); layer-wise analysis further shows that residual class information persists across network depths. These findings call for representation-aware evaluation standards in federated unlearning research. Code is publicly available at https://github.com/YuZhenyuLindy/Mirage.
  
\end{abstract}

\section{Introduction}
\label{sec:intro}

Modern vision models are increasingly deployed in collaborative and regulated environments where selective forgetting is required~\cite{papyan2020prevalence,thudi2022necessity,yu2025forgetme,meng2026tri}. In medical imaging consortia, cross-organization feature pipelines, and large-scale model serving systems, intermediate representations are frequently shared, stored, or reused. Legal requirements such as the right to be forgotten have therefore motivated \emph{visual unlearning}, which ensures that a trained model behaves as if designated classes or samples had never been observed~\cite{yu2025instantforget}. Current evaluations focus almost exclusively on output behavior, including retained-data accuracy, forgotten-label accuracy, and computational overhead~\cite{jia2023model,chen2023boundary,zhang2024verification,kurmanji2023towards,foster2024fast,yu2025instantforget}. If the model produces uninformative predictions on the forgotten target while maintaining utility, forgetting is deemed complete.

This evaluation protocol assumes that suppressing classifier predictions implies erasing the underlying information~\cite{thudi2022necessity}. For vision models, this assumption is problematic. Deep networks encode class identity in high-dimensional feature geometries~\cite{papyan2020prevalence} where class-conditional subspaces remain separable across training variations. Modifying the classifier head does not guarantee that these internal structures collapse. A model may therefore satisfy behavioral forgetting criteria while preserving linearly recoverable class information in its intermediate embeddings~\cite{hayes2025inexact,zhang2024verification,kim2026trulyforgetting, jeon2026idi, jang2026suppression}.

The distinction becomes critical in settings such as Vertical Federated Learning (VFL)~\cite{vepakomma2018split}, where multiple passive parties compute and transmit intermediate embeddings to an active party holding labels. In this scenario, representations are explicitly exposed and can be analyzed independently of the classifier head. Yet existing vertical federated unlearning methods largely inherit output-level certification from federated learning literature~\cite{che2023fast,wang2024efficient,varshney2025unlearning}, leaving the internal representation space effectively unaudited.

To examine whether visual models truly forget at the representational level, in this paper we introduce \textbf{Mirage}, a post-hoc auditing framework that probes recoverability, structural alignment, feature separability, and layer-wise information persistence. Mirage evaluates unlearning relative to a retrained-from-scratch baseline and formalizes the \emph{forgetting gap}, which measures excess recoverability beyond what intrinsic class geometry would induce. This retraining-relative perspective enables principled certification of representation-level erasure rather than mere output suppression. 
Our main \textbf{contributions} are:
\begin{itemize}
    \item We introduce \textbf{Mirage}, a representation-level auditing framework that evaluates visual unlearning through complementary analyses of recoverability, structural alignment, and layer-wise information persistence.
    
    \item We formalize the \textbf{forgetting gap}, a retraining-relative criterion that quantifies residual representation structure beyond intrinsic class geometry.
    
    \item Through extensive empirical analysis using all four diagnostics, we reveal two systematic phenomena in visual unlearning: a misalignment between output-level forgetting and representation-level erasure, where Centered Kernel Alignment (CKA) and separability scores corroborate the Linear Probe Recovery (LPR)-based forgetting gap; and a \textbf{class-sample asymmetry}, in which class-level forgetting leaves strong representational traces, with positive gaps persisting across network depths, while sample-level forgetting is indistinguishable from random guessing.
\end{itemize}

\section{Related Work}
\label{sec:related}

\subsection{Machine Unlearning and Data Deletion}
Machine unlearning aims to remove the influence of selected training data without full retraining~\cite{cao2015towards,bourtoule2021machine}. Early work formalized the problem as efficient data deletion~\cite{ginart2019making}, while subsequent methods proposed partition-based retraining strategies such as SISA~\cite{bourtoule2021machine}. Approximate approaches include gradient-based removal~\cite{golatkar2020eternal}, influence-function analysis~\cite{koh2017understanding}, and distillation-based correction~\cite{chundawat2023zero}. A parallel line of research pursues certified unlearning through differential-privacy-style guarantees or information-theoretic bounds on the residual influence of deleted data~\cite{yang2025mic,wang2025zero}. These certified approaches provide statistical guarantees on output distributions but do not directly examine the geometry of internal representations \cite{hayes2025inexact}. Mirage complements such guarantees by auditing whether representational structure, rather than output behavior, aligns with a retrained reference.

\subsection{Federated and Distributed Unlearning}
In federated learning, unlearning addresses client removal and data deletion in distributed settings~\cite{liu2021federaser}. While most existing works focus on horizontal federated learning, collaborative paradigms such as split learning explicitly transmit intermediate representations across parties~\cite{vepakomma2018split}. In such settings, internal embeddings are shared and may be analyzed independently of the classifier head. However, evaluation protocols for federated unlearning largely inherit centralized output-level metrics, leaving the internal representation space unaudited.

\subsection{Representation Probing and Geometric Analysis}
A substantial body of work studies what neural networks encode in their intermediate layers. Linear probing has been widely used to assess recoverable information in learned representations~\cite{alain2016understanding,belinkov2022probing}. Representational similarity analysis, including Centered Kernel Alignment (CKA)~\cite{kornblith2019similarity}, provides tools for comparing internal structures across models. These studies demonstrate that classifier performance does not fully characterize the geometry of embedding spaces. In vision models, class-conditional structure often remains linearly separable even under architectural or training variations. However, such representation-level analyses have rarely been integrated into unlearning evaluation \cite{kim2026trulyforgetting, jeon2026idi, hayes2025inexact}.

\subsection{Representation Leakage and Privacy Risks}
Prior research has shown that trained models can retain sensitive information beyond their observable outputs. Membership inference attacks~\cite{shokri2017membership} and memorization analyses~\cite{song2017machine} reveal that internal representations may encode recoverable attributes even when output behavior appears constrained. In distributed learning frameworks where embeddings are transmitted or reused, this representational persistence raises additional risks. These findings suggest that certifying unlearning requires examining the recoverability and structural properties of internal representations rather than relying solely on output-level metrics.

\section{Preliminaries}
\label{sec:prelim}

\subsection{Collaborative Visual Learning Setting}
\label{sec:vfl}

We consider a collaborative visual learning setting in which feature extraction and classification are decoupled across parties. A model consists of $K$ bottom encoders $G_{\theta_k}$ and a top classifier $F_\omega$. For input sample $x = (x_1, \ldots, x_K)$, each party computes an embedding
\begin{equation}
H_k = G_{\theta_k}(x_k),
\end{equation}
and the top model predicts
\begin{equation}
\hat{y} = F_\omega(H_1, \ldots, H_K).
\end{equation}
The joint parameter set $\Theta = (\theta_1, \ldots, \theta_K, \omega)$ minimizes
\begin{equation}
\min_{\Theta} \frac{1}{n} \sum_{i=1}^{n}
\ell\bigl(F_\omega(G_{\theta_1}(x_{1,i}), \ldots, G_{\theta_K}(x_{K,i})), y_i\bigr).
\label{eq:vfl-obj}
\end{equation}

Vertical Federated Learning (VFL) is a representative instance of this formulation, where one active party holds labels and passive parties hold disjoint feature partitions. The key property of this setting is that intermediate representations $H_k$ are explicitly exposed across parties.

\subsection{Label Unlearning Objective}
\label{sec:label-unlearning}

Given a trained model $\Theta^*$ and a target label set $\mathcal{Y}_u$ that are to be unlearned, label unlearning seeks to remove all information associated with $\mathcal{Y}_u$. Let
\begin{equation}
    D_u = \{(x_i, y_i) \mid y_i \in \mathcal{Y}_u\},
    \qquad
    D_r = D \setminus D_u.
\end{equation}

An unlearned model $\Theta^u$ is considered ideal if it is functionally indistinguishable from a model $\Theta^r$ retrained from scratch on $D_r$. 
Existing evaluation protocols rely on output-level metrics. Retained accuracy ($\text{Acc}_r$) measures performance on $D_r$. Forgotten-label accuracy ($y_u$) measures prediction accuracy on $D_u$, where lower values indicate stronger behavioral forgetting. Runtime measures computational efficiency. These metrics assess classifier behavior but do not directly examine the internal representation space.

\subsection{Representation-Level Criterion}
\label{sec:hypothesis}

Let $\phi_l(\cdot)$ denote the feature map of layer $l$. For a model $\Theta$, define its linear probe recoverability on forgotten data as
\begin{equation}
\text{LPR}(\Theta)
=
\max_{h \in \mathcal{H}_{\text{lin}}}
\mathbb{E}_{x \in D}
\bigl[
\mathbf{1}[h(\phi_l(x)) = \mathbf{1}[y \in \mathcal{Y}_u]]
\bigr],
\label{eq:lpr}
\end{equation}
where $\mathcal{H}_{\text{lin}}$ is the set of linear classifiers.
We define the \emph{forgetting gap} as
\begin{equation}
\Delta_{\text{LPR}}
=
\text{LPR}(\Theta^u)
-
\text{LPR}(\Theta^r).
\label{eq:fg}
\end{equation}

The retrained baseline $\Theta^r$ captures intrinsic class structure that naturally emerges in visual representations. A positive forgetting gap indicates residual class-specific structure retained by the unlearned model beyond what retraining would preserve. This criterion provides a representation-level perspective on whether unlearning achieves erasure rather than suppression of output behavior.

\begin{figure}[t]
    \centering
    \includegraphics[width=1\linewidth]{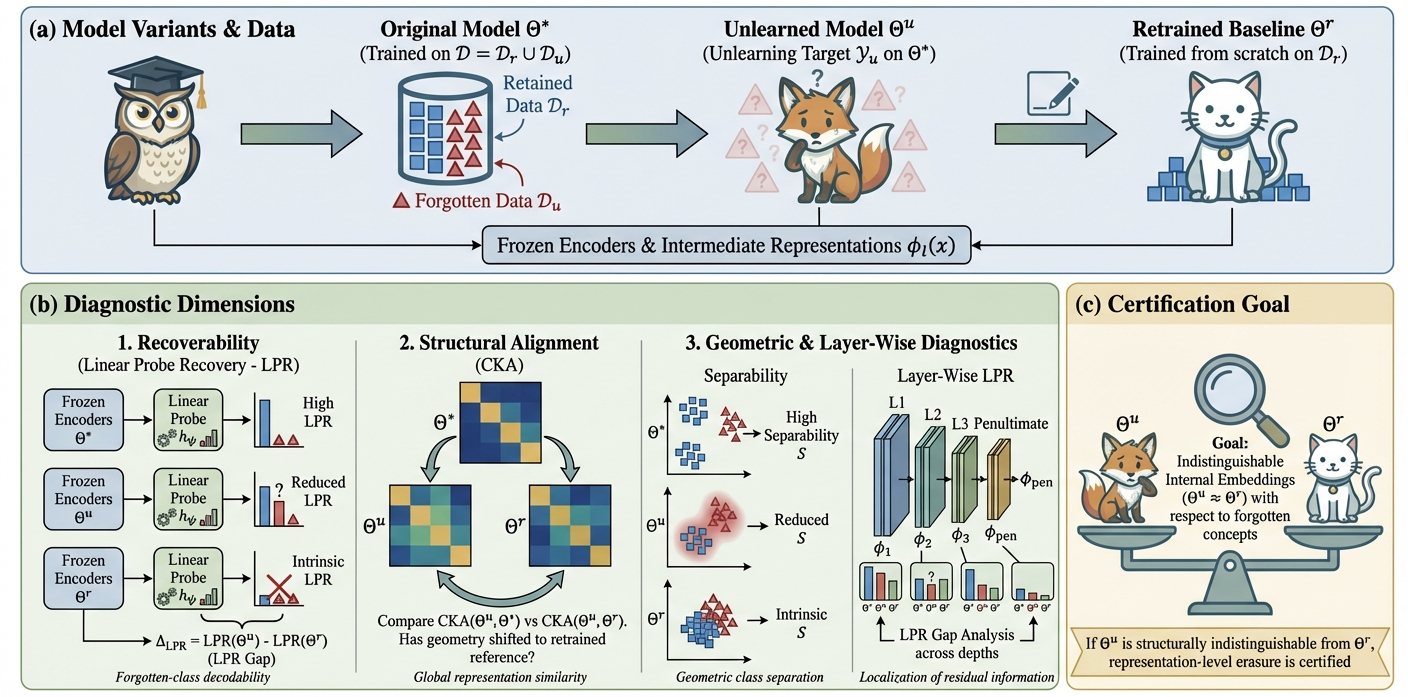}
    \caption{\textbf{Mirage: a representation-level certification framework.} Mirage audits representation-level forgetting by comparing the original model $\Theta^*$, the unlearned model $\Theta^u$, and a retrained baseline $\Theta^r$ trained from scratch on retained data. The framework analyzes Linear Probe Recovery (LPR), Centered Kernel Alignment (CKA), feature separability, and layer-wise information persistence. Representation-level erasure is certified when the internal embeddings of $\Theta^u$ become structurally indistinguishable from the retrained reference.}
    \label{fig:overview}
\end{figure}

\section{Methodology: the Mirage Framework}
\label{sec:method}

Mirage provides a principled framework for certifying representation-level forgetting in visual models (see \Cref{fig:overview}). The central question is whether unlearning removes class-specific structure from embeddings or suppresses the classifier head. Mirage answers this through complementary analyses of recoverability, geometric alignment, and spatial localization, all evaluated relative to a retrained-from-scratch baseline that defines the reference representation geometry.

\subsection{Certification Objective}
\label{sec:cert-objective}

Let $\Theta^*$ denote the original trained model, $\Theta^u$ the unlearned model, and $\Theta^r$ a model retrained from scratch on retained data $D_r$. Representation-level certification aims to determine whether $\Theta^u$ is indistinguishable from $\Theta^r$ with respect to forgotten-label information in its internal embeddings.

\noindent Certification therefore requires that, for each diagnostic measure $\mathcal{D}$ in Mirage,

\begin{equation}
|\mathcal{D}(\Theta^u) - \mathcal{D}(\Theta^r)| \le \epsilon,
\end{equation}
under matched evaluation protocols. Any systematic deviation indicates residual representational traces beyond what retraining would preserve. Mirage evaluates this condition through three complementary dimensions: recoverability, structural alignment, and geometric localization.

\subsection{Recoverability and Structural Analysis}
\label{sec:recoverability}

This component examines whether forgotten-label information remains decodable and global embedding geometry has shifted toward the retrained reference.

\subsubsection{Linear Probe Recovery.}
\label{sec:lpr}

Linear Probe Recovery (LPR) evaluates whether forgotten-label information remains linearly decodable from frozen embeddings. The formal definition of LPR and the forgetting gap $\Delta_{\text{LPR}}$ are given in Eqs. (\ref{eq:lpr})--(\ref{eq:fg}).
Concretely, given an unlearned model $\Theta^u$, we freeze its parameters and extract features from layer $l$, denoted by $\phi_l(x) \in \mathbb{R}^{d_l}$. A linear classifier $h_\psi$ is trained via cross-entropy to predict membership in the forgotten label set $\mathcal{Y}_u$:
\begin{equation}
\min_\psi
\frac{1}{|D_{\text{probe}}|}
\sum_{(x_i, y_i) \in D_{\text{probe}}}
\ell_{\text{CE}}\bigl(h_\psi(\phi_l(x_i)),\; \mathbf{1}[y_i \in \mathcal{Y}_u]\bigr).
\label{eq:lpr-opt}
\end{equation}

The probe thus performs binary classification by distinguishing samples belonging to $\mathcal{Y}_u$ from retained samples. Because visual representations naturally encode class structure, absolute LPR alone is insufficient. Mirage therefore evaluates $\Delta_{\text{LPR}}$ relative to the retrained baseline. A positive $\Delta_{\text{LPR}}$ indicates residual class-specific information beyond intrinsic representation geometry.

\paragraph{Recoverability Interpretation.}

If forgotten-class features remain separated from retained features in embedding space, linear classification accuracy exceeds chance. Under equal class priors and isotropic Gaussian class-conditional assumptions, let the signal-to-noise ratio be defined as

\begin{equation}
\text{SNR}
=
\frac{\|\mu_u - \mu_r\|^2}{\sigma^2},
\end{equation}
the optimal linear classifier satisfies

\begin{equation}
\text{Acc}_{\text{LPR}}
\ge
\Phi\!\left(\frac{\sqrt{\text{SNR}}}{2}\right),
\end{equation}
where $\Phi$ is the standard normal cumulative distribution function. Under the isotropic assumption ($\Sigma_u = \Sigma_r = \sigma^2 I$), the feature separability score $\mathcal{F}$ defined in \Cref{sec:separability} satisfies $\mathcal{F} = \text{SNR}/(2d)$, so $\mathcal{F}$ is a dimension-normalized proxy for SNR that preserves the same monotone relationship with probe accuracy. Empirically, methods with high separability in \Cref{tab:mirage-main} (e.g., FT on COVID-19: $\mathcal{F} = 0.280$) indeed exhibit high LPR ($96.2\%$), while methods with separability close to the retrained level (e.g., UNSIR on COVID-19: $\mathcal{F} = 0.040$ vs.\ Retrain $\mathcal{F} = 0.031$) show LPR near the baseline. This confirms that feature-space geometry governs linear recoverability, and $\mathcal{F}$ provides a principled, probe-independent measure of residual class structure.

\subsubsection{CKA-Based Structural Alignment.}
\label{sec:cka}

While LPR measures decodability, it does not capture global geometric structure. Mirage therefore employs Centered Kernel Alignment (CKA)~\cite{kornblith2019similarity}. 
For any two models $\Theta_a$ and $\Theta_b$, let $X \in \mathbb{R}^{n \times p}$ and $Y \in \mathbb{R}^{n \times q}$ denote their respective layer-$l$ representations computed on the same $n$ inputs. Linear CKA is defined as

\begin{equation}
\text{CKA}(X, Y)
=
\frac{\|Y^\top X\|_F^2}
{\|X^\top X\|_F \cdot \|Y^\top Y\|_F}.
\end{equation}

We compare representations among $\Theta^*$, $\Theta^u$, and $\Theta^r$. If $\Theta^u$ remains more aligned with $\Theta^*$ than with $\Theta^r$, the unlearning procedure has not reshaped the embedding geometry toward the retrained reference. In such cases, behavioral forgetting may coexist with structural persistence.

\subsection{Geometric and Layer-Wise Diagnostics}
\label{sec:geometry}

This component examines distribution-level separation and the spatial distribution of residual information across network depth.

\subsubsection{Feature Separability.}
\label{sec:separability}

To quantify geometric class separation independent of probe optimization, Mirage computes a Fisher-inspired separability score. Let $\mu_u$ and $\Sigma_u$ denote the mean and covariance of embeddings from the forgotten set $D_u$, and $\mu_r$ and $\Sigma_r$ denote those from the retained set $D_r$. The score is:

\begin{equation}
\mathcal{F}
=
\frac{\|\mu_u - \mu_r\|^2}
{\mathrm{tr}(\Sigma_u) + \mathrm{tr}(\Sigma_r)}.
\label{eq:sep}
\end{equation}

A separability value exceeding that of the retrained baseline indicates residual structural discrimination of forgotten classes.

\subsubsection{Layer-Wise Recovery.}
\label{sec:layerwise}

Residual information may localize to specific network depths. Mirage evaluates LPR across early, intermediate, and penultimate layers:

\begin{equation}
\Delta_{\text{LPR}}^{(l)}
=
\text{LPR}_l(\Theta^u)
-
\text{LPR}_l(\Theta^r),
\end{equation}
where $\text{LPR}_l$ denotes recoverability measured at layer $l$. Persistent positive gaps across multiple depths indicate that unlearning has not altered representation hierarchies beyond superficial classifier adjustments (see Appendix~\ref{sec:appendix-layerwise}).

\section{Experiments}
\label{sec:exp}

\subsection{Experimental Setup}
\label{sec:settings}

\paragraph{Datasets.}
We evaluate on seven datasets:
\textbf{MNIST}~\cite{lecun2002gradient}, 
\textbf{CIFAR-10}~\cite{krizhevsky2009learning}, 
\textbf{CIFAR-100}~\cite{krizhevsky2009learning}, 
\textbf{ModelNet}~\cite{wu20153d}, 
\textbf{Brain Tumor MRI}~\cite{wang2024efficient}, 
\textbf{COVID-19 Radiography}~\cite{chowdhury2020can,rahman2021exploring}, 
and \textbf{Yahoo Answers}~\cite{chen2020mixtext}. The first six datasets are image classification benchmarks with increasing visual complexity, while Yahoo Answers is a text classification dataset included to verify that the proposed evaluation protocol is not limited to purely visual inputs.

\paragraph{Baseline Methods.}
We compare the following unlearning methods: \textbf{Retrain} (retraining from scratch on $D_r$), \textbf{Fine-Tuning} (FT)~\cite{golatkar2020eternal}, \textbf{Fisher Forgetting}~\cite{golatkar2020eternal}, \textbf{Amnesiac Unlearning}~\cite{graves2021amnesiac}, \textbf{UNSIR}~\cite{tarun2023fast}, \textbf{Boundary Unlearning} (BU)~\cite{chen2023boundary}, \textbf{SSD}~\cite{foster2024fast}, and the \textbf{Target Method}~\cite{gutowards2026}. All implementations follow the original specifications and hyperparameters provided by the respective authors.

\paragraph{Evaluation Metrics.}
\textbf{Output-Level Metrics.}
We report retained accuracy $\text{Acc}_r$ and forgotten-label accuracy $y_u$ as defined in \Cref{sec:label-unlearning}. Effective output-level forgetting requires low $y_u$ while maintaining high $\text{Acc}_r$.
\textbf{Representation-Level Metrics.}
We apply the Mirage diagnostics described in Linear Probe Recovery (LPR) and its gap $\Delta_{\text{LPR}}$ (\Cref{sec:lpr}), CKA-based structural alignment (\Cref{sec:cka}), and feature separability (\Cref{sec:separability}). For LPR, we use a logistic regression probe with $\ell_2$ regularization ($C=1.0$), trained for 1000 iterations on features extracted from the penultimate layer. CKA is computed using linear CKA on 5000 randomly sampled data points. Feature separability uses all available data from $D_u$ and a balanced sample from $D_r$.

\paragraph{Implementation Details.}
We follow the experimental protocol~\cite{gutowards2026}, including identical data splits, model architectures, optimization settings, and unlearning procedures to ensure fair comparison. For all vision datasets, we adopt \textbf{ResNet18}~\cite{he2016deep} as the backbone model, while for Yahoo Answers we use a multi-layer perceptron (MLP) consistent with the baseline configuration. In the vertical federated learning (VFL) setting, input features are partitioned equally between two passive parties, while the active party holds the labels and the top classifier. All models are first trained on the full dataset $D$ to obtain the original model $\Theta^*$. For single-label unlearning, one class $\mathcal{Y}_u$ is designated as the forgotten class and the remaining data form the retained set $D_r$. For sample-level unlearning, 5\% or 10\% of samples are randomly selected for removal. The retrained baseline $\Theta^r$ is obtained by training from scratch on $D_r$ with the same architecture and hyperparameters. Results are averaged over three random seeds.

\subsection{Comparison}
\label{sec:comparison}

\subsubsection{Quantitative Analysis}
\label{sec:quant}

We begin with quantitative comparison across all seven datasets using both output-level and representation-level metrics.

\paragraph{Output-Level Behavior.}
As shown in \Cref{tab:output-level}, no method consistently achieves both high $\text{Acc}_r$ and low $y_u$. FT and SSD maintain $\text{Acc}_r$ close to Retrain but exhibit high $y_u$, which indicates ineffective forgetting. Target and Fisher reduce $y_u$ to zero on most datasets, but this reduction is accompanied by severe utility degradation. Amnesiac shows unstable behavior across datasets. BU is the only method that achieves near-zero $y_u$ while preserving reasonable $\text{Acc}_r$ on selected datasets such as Brain Tumor, COVID-19, and Yahoo Answers.

\begin{table}[t]
\centering
\caption{Output-level metrics for single-label unlearning across seven datasets. $\text{Acc}_r$ (\%) is retained accuracy ($\uparrow$) and $y_u$ (\%) is forgotten-label accuracy ($\downarrow$). \textbf{Bold} entries indicate apparent output-level success.}
\label{tab:output-level}
\footnotesize
\resizebox{\linewidth}{!}{
\begin{tabular}{l cc cc cc cc cc cc cc}
\toprule
\multirow{2}{*}{\textbf{Method}} & \multicolumn{2}{c}{\textbf{MNIST}} & \multicolumn{2}{c}{\textbf{CIFAR-10}} & \multicolumn{2}{c}{\textbf{CIFAR-100}} & \multicolumn{2}{c}{\textbf{Brain Tumor}} & \multicolumn{2}{c}{\textbf{COVID-19}} & \multicolumn{2}{c}{\textbf{ModelNet}} & \multicolumn{2}{c}{\textbf{Yahoo Answers}} \\
\cmidrule(lr){2-3} \cmidrule(lr){4-5} \cmidrule(lr){6-7} \cmidrule(lr){8-9} \cmidrule(lr){10-11} \cmidrule(lr){12-13} \cmidrule(lr){14-15}
~ & Acc$_r$ & $y_u$ & Acc$_r$ & $y_u$ & Acc$_r$ & $y_u$ & Acc$_r$ & $y_u$ & Acc$_r$ & $y_u$ & Acc$_r$ & $y_u$ & Acc$_r$ & $y_u$ \\
\midrule
\rowcolor{tablerowalt}
Retrain & 99.4 & 0.0 & 89.7 & 0.0 & 62.2 & 0.0 & 99.1 & 0.0 & 92.8 & 0.0 & 82.3 & 0.0 & 56.7 & 0.0 \\
FT      & 99.3 & 99.4 & 89.7 & 48.5 & 61.9 & 31.7 & 98.5 & 77.4 & 92.0 & 91.8 & 81.7 & 52.7 & 56.4 & 5.0 \\
\rowcolor{tablerowalt}
Fisher  & 6.8 & 33.3 & 9.0 & 14.6 & 0.9 & 0.0 & 13.7 & 58.9 & 4.6 & 64.4 & 11.1 & 0.0 & 11.4 & 0.1 \\
Amnesiac & 75.0 & 12.6 & 70.1 & 13.5 & 30.1 & 0.0 & 69.2 & 19.6 & 57.5 & 24.1 & 75.3 & 29.3 & 42.5 & 2.7 \\
\rowcolor{tablerowalt}
UNSIR   & 48.4 & 0.0 & 23.3 & 0.0 & 2.4 & 0.0 & 51.4 & 0.0 & \textbf{70.9} & \textbf{0.0} & 14.9 & 5.3 & 32.7 & 0.0 \\
BU      & 44.7 & 0.0 & 47.4 & 0.0 & 33.4 & 0.0 & \textbf{72.4} & \textbf{0.0} & \textbf{77.9} & \textbf{0.0} & 73.3 & 8.7 & 51.7 & 0.1 \\
\rowcolor{tablerowalt}
SSD     & 99.3 & 97.0 & 89.6 & 88.2 & 62.7 & 72.7 & 98.8 & 74.7 & 92.1 & 90.9 & 81.7 & 39.3 & 55.2 & 37.6 \\
Target  & 14.9 & 0.0 & 24.1 & 0.0 & 1.0 & 0.0 & 33.3 & 0.0 & 34.3 & 0.0 & 11.8 & 0.0 & 49.7 & 0.0 \\
\bottomrule
\end{tabular}
}
\end{table}

\paragraph{Representation-Level Diagnostics.}
Mirage reveals a starkly different picture. As shown in \Cref{tab:mirage-main}, substantial LPR values ($54$ to $83\%$ across datasets) are observed even for the Retrain model, confirming that class structure is inherently encoded in visual representations. Absolute LPR values are therefore uninformative; the forgetting gap $\Delta_{\text{LPR}}$ relative to Retrain is the principled criterion.

BU exhibits consistently positive $\Delta_{\text{LPR}}$ on datasets where it appears successful at the output level. On COVID-19, BU achieves $y_u = 0\%$ while producing an LPR of $94.7\%$, which is $15.4$ points above the Retrain baseline. This indicates that forgotten-class information remains linearly recoverable in feature space.

FT and SSD show even larger positive $\Delta_{\text{LPR}}$, which aligns with their high CKA similarity to the original model. Target shows negative $\Delta_{\text{LPR}}$ on several datasets. However, this coincides with catastrophic utility collapse and low CKA values, suggesting that representational structure is destroyed rather than selectively modified. The only exception is Yahoo Answers, where Target retains high CKA ($0.957$) and a positive $\Delta_{\text{LPR}}$ ($+7.8$), behaving like a suppression-style method rather than collapsing.

\paragraph{Feature Separability.}

\begin{wrapfigure}{r}{0.48\textwidth}
\vspace{-25pt}
  \centering
  \includegraphics[width=\linewidth]{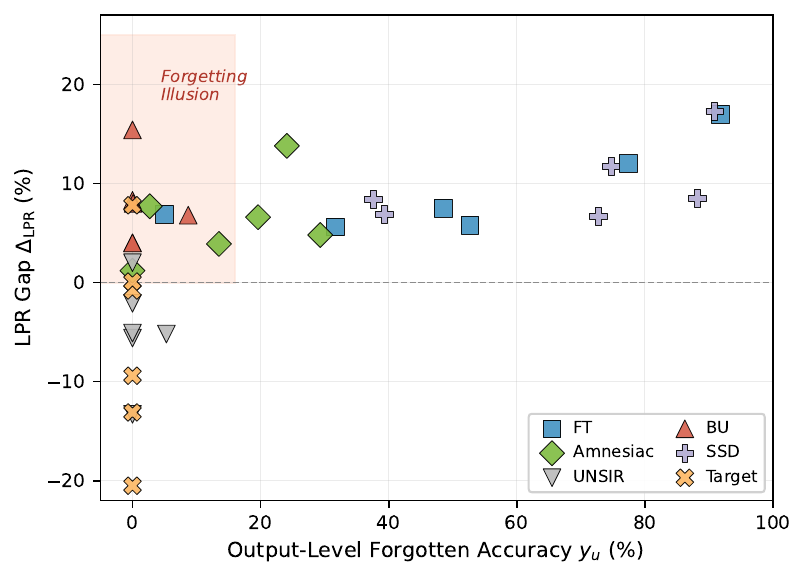}
  \vspace{-20pt}
  \caption{Forgetting gap across methods and datasets. Each point represents a method--dataset pair with coordinates $(y_u, \Delta_{\text{LPR}})$. The red region ($y_u \approx 0$, $\Delta_{\text{LPR}} > 0$) is the \emph{forgetting illusion}. BU (triangles) consistently falls here.}
  \label{fig:forgetting-gap}
  \vspace{-20pt}
\end{wrapfigure}

The separability scores in \Cref{tab:mirage-main} provide a complementary perspective independent of probe optimization. FT and SSD exhibit the highest separability values (e.g., $0.280$ and $0.340$ on COVID-19), reflecting their near-complete preservation of feature geometry. BU shows moderate separability, lower than FT and SSD but consistently above Retrain. This confirms that the forgotten class remains geometrically distinguishable. Target produces heterogeneous separability. It is near-zero on datasets where it collapses utility (e.g., $0.038$ on CIFAR-100), but occasionally high (e.g., $0.354$ on Brain Tumor) where residual class structure persists in a distorted feature space. Separability and LPR are correlated but not redundant. High separability generally accompanies high LPR, yet cases like UNSIR on CIFAR-10 ($\text{Sep} = 0.097$, $\text{LPR} = 69.5\%$) show that moderate linear decodability can occur even with low geometric separation, suggesting the probe exploits subtle directional structure beyond centroid differences. These patterns are shown in \Cref{fig:forgetting-gap}, where BU consistently lies in the forgetting-illusion region ($y_u \approx 0$, $\Delta_{\text{LPR}} > 0$) while Target falls into a collapse regime with degraded utility on the vision datasets. Output-level forgetting therefore does not imply representation-level erasure, and Mirage distinguishes these two failure modes, revealing cases where apparent forgetting masks substantial residual class structure.

\begin{table}[t]
\centering
\caption{Mirage representation-level diagnostics for single-label unlearning. LPR denotes Linear Probe Recovery (\%) and $\Delta$ denotes the LPR gap relative to Retrain. CKA$_O$ measures similarity to the original model and Sep denotes feature separability. Red values indicate $\Delta_{\text{LPR}} > 5\%$. \textbf{MNIST} is omitted because all methods achieve near-ceiling LPR ($>97\%$) with negligible $\Delta_{\text{LPR}}$ ($<1.5$ pp), making gap analysis uninformative (output-level results remain in Table~\ref{tab:output-level}). \textbf{Fisher} is omitted because it causes catastrophic utility collapse ($\text{Acc}_r < 14\%$ on all vision datasets, see Table~\ref{tab:output-level}), rendering representation diagnostics uninformative.}
\label{tab:mirage-main}
\resizebox{1.0\linewidth}{!}{
\setlength{\tabcolsep}{5pt}
\begin{tabular}{l cccc cccc cccc}
\toprule
\multirow{2}{*}{\textbf{Method}} & \multicolumn{4}{c}{\textbf{CIFAR-10}} & \multicolumn{4}{c}{\textbf{COVID-19}} & \multicolumn{4}{c}{\textbf{Brain Tumor}} \\
\cmidrule(lr){2-5} \cmidrule(lr){6-9} \cmidrule(lr){10-13}
~ & LPR & $\Delta$ & CKA$_O$ & Sep & LPR & $\Delta$ & CKA$_O$ & Sep & LPR & $\Delta$ & CKA$_O$ & Sep \\
\midrule
\rowcolor{tablerowalt}
Retrain & 82.8 & --- & .989 & .093 & 79.2 & --- & .981 & .031 & 79.8 & --- & .978 & .176 \\
FT      & 90.4 & \textcolor{forgettinggap}{+7.5} & .999 & .160 & 96.2 & \textcolor{forgettinggap}{+17.0} & .998 & .280 & 91.8 & \textcolor{forgettinggap}{+12.0} & .986 & .549 \\
\rowcolor{tablerowalt}
Amnesiac & 86.7 & +3.9 & .974 & .103 & 93.0 & \textcolor{forgettinggap}{+13.8} & .971 & .091 & 86.4 & \textcolor{forgettinggap}{+6.6} & .943 & .234 \\
UNSIR   & 69.5 & $-$13.3 & .947 & .097 & 77.2 & $-$2.1 & .954 & .040 & 74.2 & $-$5.6 & .936 & .118 \\
\rowcolor{tablerowalt}
BU      & 86.8 & +4.0 & .956 & .149 & 94.7 & \textcolor{forgettinggap}{\textbf{+15.4}} & .957 & .115 & 88.1 & \textcolor{forgettinggap}{\textbf{+8.3}} & .909 & .289 \\
SSD     & 91.3 & \textcolor{forgettinggap}{+8.5} & 1.00 & .246 & 96.5 & \textcolor{forgettinggap}{+17.3} & 1.00 & .340 & 91.5 & \textcolor{forgettinggap}{+11.7} & 1.00 & .641 \\
\rowcolor{tablerowalt}
Target  & 82.0 & $-$0.8 & .928 & .104 & 69.9 & $-$9.4 & .841 & .089 & 79.9 & +0.1 & .816 & .354 \\
\bottomrule
\\[-6pt]
\toprule
\multirow{2}{*}{\textbf{Method}} & \multicolumn{4}{c}{\textbf{CIFAR-100}} & \multicolumn{4}{c}{\textbf{ModelNet}} & \multicolumn{4}{c}{\textbf{Yahoo Answers}} \\
\cmidrule(lr){2-5} \cmidrule(lr){6-9} \cmidrule(lr){10-13}
~ & LPR & $\Delta$ & CKA$_O$ & Sep & LPR & $\Delta$ & CKA$_O$ & Sep & LPR & $\Delta$ & CKA$_O$ & Sep \\
\midrule
\rowcolor{tablerowalt}
Retrain & 73.2 & --- & .992 & .453 & 72.4 & --- & .990 & .132 & 53.7 & --- & .926 & .008 \\
FT      & 78.8 & \textcolor{forgettinggap}{+5.6} & .999 & .641 & 78.2 & \textcolor{forgettinggap}{+5.8} & 1.00 & .241 & 60.6 & \textcolor{forgettinggap}{+6.9} & .992 & .020 \\
\rowcolor{tablerowalt}
Amnesiac & 74.4 & +1.2 & .986 & .284 & 77.2 & +4.8 & .983 & .191 & 61.4 & \textcolor{forgettinggap}{+7.7} & .927 & .020 \\
UNSIR   & 68.1 & $-$5.1 & .960 & .277 & 67.2 & $-$5.2 & .921 & .030 & 55.7 & +2.0 & .836 & .003 \\
\rowcolor{tablerowalt}
BU      & 77.1 & +4.0 & .985 & .381 & 79.2 & \textcolor{forgettinggap}{+6.8} & .969 & .165 & 61.7 & \textcolor{forgettinggap}{+8.0} & .939 & .018 \\
SSD     & 79.9 & \textcolor{forgettinggap}{+6.7} & 1.00 & .686 & 79.3 & \textcolor{forgettinggap}{+6.9} & 1.00 & .240 & 62.1 & \textcolor{forgettinggap}{+8.4} & .999 & .023 \\
\rowcolor{tablerowalt}
Target  & 52.7 & $-$20.5 & .622 & .038 & 59.4 & $-$13.1 & .778 & .021 & 61.5 & \textcolor{forgettinggap}{+7.8} & .957 & .020 \\
\bottomrule
\end{tabular}
}
\end{table}

\subsubsection{Qualitative Analysis}
\label{sec:qual}

We examine the geometry of learned representations through feature-space visualization and structural comparison.

\begin{figure}[t]
  \centering
  \includegraphics[width=\linewidth]{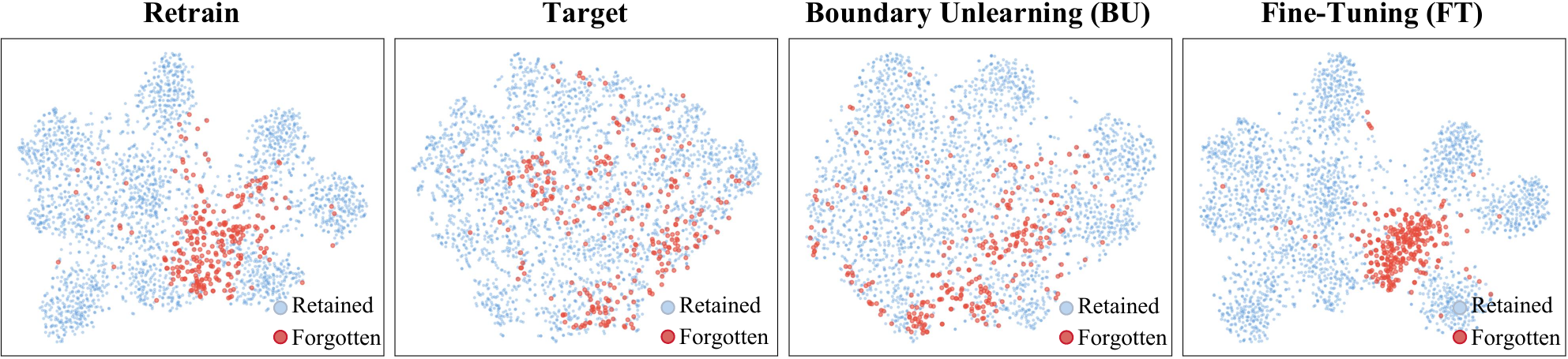}
  \caption{t-SNE visualization of bottom-model features on COVID-19. Blue: retained classes; red: forgotten class. \textbf{Retrain} (left) shows the forgotten class forming a separable cluster even without training on its labels. \textbf{Target} scatters all points, reflecting model collapse ($\text{Acc}_r = 34.3\%$). \textbf{BU} preserves the forgotten-class cluster almost identically to Retrain, visually confirming the forgetting illusion ($\Delta_{\text{LPR}} = +15.4\%$). \textbf{FT} likewise preserves the full representational structure.}
  \label{fig:tsne}
\end{figure}

\paragraph{t-SNE Visualization.}
t-SNE embeddings of bottom-model features on the COVID-19 dataset are presented in \Cref{fig:tsne}. In the Retrain model, the forgotten class forms a clearly separable cluster despite the absence of its labels during training. This confirms that general-purpose representations naturally encode class-level structure. For BU, the forgotten-class cluster remains compact and well-separated from retained classes, and its spatial configuration closely resembles that of Retrain. Although output-level accuracy on the forgotten class drops to zero, the geometric structure persists. This visual evidence is consistent with the large positive $\Delta_{\text{LPR}}$ observed quantitatively. In contrast, Target produces a scattered feature distribution with no coherent clustering. The collapse of structure aligns with the sharp drop in retained accuracy and the low CKA similarity. The apparent reduction in LPR is therefore attributable to representational destruction rather than targeted forgetting. Analogous patterns are observed across all remaining datasets (see \Cref{fig:tsne-mnist,fig:tsne-cifar10,fig:tsne-cifar100,fig:tsne-braintumor,fig:tsne-modelnet,fig:tsne-yahoo} in the Appendix).

\paragraph{Representation Similarity Patterns.}
We further compare CKA similarity across methods. FT and SSD exhibit near-unity CKA with the original model, indicating minimal structural change. BU shows moderately high similarity, which suggests that most of the representation space remains intact. Target shows substantially lower similarity on several datasets, reflecting large-scale alteration of feature geometry. These qualitative patterns reinforce the quantitative findings by demonstrating that methods appearing successful under output-level evaluation can fully preserve the geometric signature of the forgotten class. Mirage makes this residual structure directly visible and quantifiable.

\subsubsection{Class-Sample Asymmetry}
\label{sec:sample-results}

We extend the analysis to sample-level unlearning, where a fixed proportion (5\% or 10\%) of randomly selected samples is designated for forgetting rather than entire classes. The LPR results are shown in \Cref{tab:sample-level}.


\begin{table}[t]
\centering
\caption{Sample-level unlearning results across seven datasets. LPR (\%) for forgetting 5\% and 10\% of random samples (mean over 3 rounds). All methods, including Retrain, remain near chance level ($\approx 50\%$), in contrast to class-level recovery (Table~\ref{tab:mirage-main}).}
\label{tab:sample-level}
\resizebox{\linewidth}{!}{
\begin{tabular}{l cc cc cc cc cc cc cc}
\toprule
\multirow{2}{*}{\textbf{Method}} & \multicolumn{2}{c}{\textbf{MNIST}} & \multicolumn{2}{c}{\textbf{CIFAR-10}} & \multicolumn{2}{c}{\textbf{CIFAR-100}} & \multicolumn{2}{c}{\textbf{Brain Tumor}} & \multicolumn{2}{c}{\textbf{COVID-19}} & \multicolumn{2}{c}{\textbf{ModelNet}} & \multicolumn{2}{c}{\textbf{Yahoo Answers}} \\
\cmidrule(lr){2-3} \cmidrule(lr){4-5} \cmidrule(lr){6-7} \cmidrule(lr){8-9} \cmidrule(lr){10-11} \cmidrule(lr){12-13} \cmidrule(lr){14-15}
 & 5\% & 10\% & 5\% & 10\% & 5\% & 10\% & 5\% & 10\% & 5\% & 10\% & 5\% & 10\% & 5\% & 10\% \\
\midrule
\rowcolor{tablerowalt}
Retrain  & 48.6 & 49.6 & 50.8 & 50.9 & 49.0 & 50.0 & 49.6 & 51.8 & 51.4 & 49.8 & 51.0 & 50.4 & 52.0 & 52.4 \\
FT       & 49.9 & 49.2 & 49.2 & 49.6 & 50.8 & 50.2 & 50.4 & 49.0 & 49.6 & 50.4 & 53.3 & 50.0 & 49.6 & 50.4 \\
\rowcolor{tablerowalt}
Amnesiac & 50.4 & 49.6 & 49.5 & 50.4 & 51.0 & 49.9 & 48.8 & 50.1 & 52.5 & 51.4 & 51.8 & 46.8 & 49.0 & 51.1 \\
UNSIR    & 49.1 & 50.2 & 49.6 & 49.5 & 50.7 & 50.4 & 52.0 & 49.6 & 51.9 & 50.2 & 44.6 & 49.6 & 51.0 & 49.2 \\
\rowcolor{tablerowalt}
BU       & 50.0 & 49.7 & 49.2 & 48.5 & 50.5 & 49.4 & 50.4 & 49.6 & 50.0 & 51.7 & 49.6 & 49.6 & 49.2 & 51.1 \\
Target   & 48.5 & 50.2 & 49.8 & 49.8 & 49.0 & 50.2 & 54.2 & 47.4 & 49.5 & 49.1 & 53.8 & 50.3 & 48.8 & 50.6 \\
\bottomrule
\end{tabular}
}
\end{table}

\noindent A striking \emph{class-sample asymmetry} emerges: class-level LPR reaches $\sim$99\% on MNIST and exceeds $96\%$ on COVID-19 (Table~\ref{tab:mirage-main}; Retrain baseline: $54$--$83\%$ on the six reported datasets), whereas sample-level LPR $\approx 50\%$ for \emph{all} methods including Retrain. This dichotomy has a clear geometric explanation: class identity is encoded in stable feature-space structures, such as cluster centers and inter-class margins, that persist across training runs. Individual sample membership, by contrast, does not manifest as a linearly separable signal in penultimate-layer embeddings. The implication is twofold: linear probes are effective and necessary for class-level auditing, but fundamentally insufficient for sample-level verification. The two forgetting settings demand categorically different diagnostic tools.

\vspace{10pt}
\subsection{Ablation Study}
\label{sec:ablation}

\subsubsection{Sensitivity to Strength.}
\label{sec:epoch-ablation}

\begin{wraptable}{r}{0.55\textwidth}
\centering
\vspace{-35pt}
\caption{Effect of unlearning epochs for BU (averaged over 3 rounds). $\Delta_{\text{LPR}}$ remains positive across epochs on all datasets.}
\label{tab:epoch-ablation}
\resizebox{0.55\textwidth}{!}{
\begin{tabular}{l ccc ccc ccc}
\toprule
& \multicolumn{3}{c}{\textbf{CIFAR-10}} & \multicolumn{3}{c}{\textbf{CIFAR-100}} & \multicolumn{3}{c}{\textbf{COVID-19}} \\
\cmidrule(lr){2-4} \cmidrule(lr){5-7} \cmidrule(lr){8-10}
\textbf{Epochs} & Acc$_r$ & LPR & $\Delta$ & Acc$_r$ & LPR & $\Delta$ & Acc$_r$ & LPR & $\Delta$ \\
\midrule
\rowcolor{tablerowalt}
0 (Ret.) & 89.7 & 83.3 & -- & 62.2 & 74.3 & -- & 92.7 & 79.5 & -- \\
1        & 79.2 & 87.7 & +4.4 & 57.4 & 80.5 & +6.2 & 89.3 & 95.4 & +15.9 \\
\rowcolor{tablerowalt}
3        & 58.9 & 87.8 & +4.5 & 42.3 & 81.6 & +7.3 & 85.6 & 94.6 & +15.1 \\
5        & 43.3 & 86.2 & +2.9 & 32.9 & 77.7 & +3.4 & 81.7 & 94.6 & +15.1 \\
\rowcolor{tablerowalt}
10       & 32.7 & 85.3 & +2.0 & 25.5 & 77.7 & +3.4 & 78.8 & 94.4 & +14.9 \\
20       & 21.8 & 83.9 & +0.6 & 22.0 & 75.4 & +1.1 & 68.2 & 95.4 & +15.9 \\
\bottomrule
\end{tabular}
}
\vspace{-25pt}
\end{wraptable}

We first analyze how increasing unlearning strength affects both utility and representation-level residuals. We vary the number of BU epochs while keeping all other hyperparameters fixed, asking: \emph{can the forgetting illusion be eliminated by simply training longer?} As shown in \Cref{tab:epoch-ablation}, the answer is no. Even at 20 epochs, $\Delta_{\text{LPR}}$ remains positive on all three datasets. 
On COVID-19, LPR stays above $94\%$ regardless of epoch count, which is $15$--$16$ points above the Retrain baseline. Meanwhile, $\text{Acc}_r$ degrades steadily, dropping on CIFAR-10 from 89.7\% (Retrain) to 21.8\% at 20 epochs. No operating point exists where representation-level residuals vanish while utility remains intact. This reveals a fundamental limitation that BU operates at the decision boundary without modifying the underlying feature geometry. More epochs progressively destroy utility, but the forgotten class remains linearly recoverable because the representational structure responsible for class separability is never altered.

\vspace{-5pt}
\subsubsection{Multi-Party Scaling.}
\label{sec:kparty-ablation}

\begin{wraptable}{r}{0.38\textwidth}
\centering
\vspace{-35pt}
\caption{Effect of the number of passive parties ($K$) on CIFAR-10. $\Delta_{\text{LPR}}$ remains positive for FT and BU but becomes negative for Target as $K$ increases.}
\label{tab:kparty}
\resizebox{\linewidth}{!}{
\begin{tabular}{l ccc ccc}
\toprule
& \multicolumn{3}{c}{\textbf{LPR (\%)}} & \multicolumn{3}{c}{\textbf{$\Delta_{\text{LPR}}$ (\%)}} \\
\cmidrule(lr){2-4} \cmidrule(lr){5-7}
\textbf{Method} & K=2 & K=4 & K=8 & K=2 & K=4 & K=8 \\
\midrule
\rowcolor{tablerowalt}
Retrain & 82.8 & 81.6 & 80.9 & -- & -- & -- \\
FT      & 90.7 & 87.6 & 87.1 & +7.9 & +6.0 & +6.2 \\
\rowcolor{tablerowalt}
BU      & 87.6 & 86.1 & 84.6 & +4.8 & +4.6 & +3.7 \\
Target  & 83.9 & 76.7 & 69.0 & +1.2 & -4.9 & -11.9 \\
\bottomrule
\end{tabular}
}
\vspace{-20pt}
\end{wraptable}

We next examine how representation-level residuals scale with the number of passive parties across which features are partitioned. We vary $K \in {2, 4, 8}$ on CIFAR-10, forgetting a single class, and measure LPR and $\Delta_{\text{LPR}}$. As shown in \Cref{tab:kparty}, BU maintains consistently positive $\Delta_{\text{LPR}}$ across all $K$. This indicates that residual information persists even as features are split across more parties. FT exhibits even larger positive gaps, consistent with its minimal representational change. In contrast, Target becomes increasingly negative as $K$ increases. This pattern reflects progressive destruction of the feature space rather than selective removal of specific class structure. The scaling behavior therefore distinguishes persistent leakage from global collapse.

\subsubsection{Sensitivity to Forgotten Class.}
\label{sec:class-ablation}

Finally, we investigate whether the forgetting gap depends on which class is selected for removal. On CIFAR-10, we test each of the 10 classes individually. The Retrain LPR varies across classes, which reflects inherent differences in class separability within the feature space. However, the relative behavior of each method remains stable. BU consistently produces positive $\Delta_{\text{LPR}}$, while Target produces values near or below zero that coincide with degraded utility. Similar patterns are observed on CIFAR-100. Although absolute LPR varies across classes, the qualitative relationship between $\Delta_{\text{LPR}}$ and retained accuracy remains unchanged. This confirms that the forgetting gap is not an artifact of a particular class choice but a structural property of the unlearning mechanism. Detailed per-class results are provided in Appendix~\ref{sec:appendix-perclass}.

\section{Discussion}
\label{sec:discussion}

\paragraph{Representation-Level Erasure as a Structural Constraint.}
Our results suggest that the gap between output-level forgetting and representation-level erasure is structural. In deep models, class identity is encoded in feature geometry through cluster structure and inter-class margins. Adjusting classifier weights or suppressing logits can change decision boundaries without altering this geometry, leaving forgotten-class information linearly recoverable. We therefore adopt a retraining-relative criterion: a retrained model reflects the geometry induced by retained data, and a positive forgetting gap ($\Delta_{\text{LPR}} > 0$) indicates residual class structure beyond retraining. Representation-level forgetting thus requires geometric alignment with the retrained reference.

\paragraph{An Empirical Unlearning Trilemma.}
Across datasets and methods, we observe a consistent tension among three desiderata: 
(1)~\textbf{utility}, 
(2)~\textbf{output forgetting}, and 
(3)~\textbf{representation forgetting} (small $\Delta_{\text{LPR}}$). 
No evaluated method achieves all three simultaneously. Methods that preserve utility often retain recoverable class structure, whereas methods that strongly suppress outputs tend to degrade retained performance. Although not a formal impossibility result, this pattern suggests a structural tension between preserving feature geometry and eliminating class separability. Additional discussion appears in Appendix~\ref{sec:appendix_discussion}.

\section{Conclusion}

We presented Mirage, a representation-level certification framework for visual unlearning. Across seven datasets, we show that suppressing forgotten-label predictions does not imply erasure of underlying feature geometry. Methods that appear successful at the output level often retain recoverable class information. By defining the forgetting gap relative to a retrained baseline, Mirage reframes unlearning evaluation as a geometric certification problem and exposes a tension among utility, behavioral forgetting, and representation-level erasure.

\section*{Protocol Alignment Statement}
We strictly follow the original datasets, architectures, unlearning scenarios, and evaluation metrics of the target baseline. Mirage introduces only representation-level diagnostics applied to the same trained models, ensuring complete experimental consistency and fair comparison.

\section*{Limitations}
Mirage is designed as an auditing framework rather than a new unlearning algorithm. Our experiments follow a VFL protocol consistent with prior work; behavior in heterogeneous real-world deployments may differ. The linear probe provides a conservative lower bound on recoverable information; nonlinear probes may reveal additional residual structure beyond linear separability (see Appendix~\ref{sec:appendix-nonlinear}). Extending certification to stronger adversaries and to horizontal federated settings remains an important open direction.


\bibliographystyle{splncs04}
\bibliography{main3}

\clearpage
\clearpage
\appendix

\setcounter{page}{1}
\setcounter{figure}{0}
\setcounter{table}{0}
\setcounter{section}{0}
\setcounter{subsection}{0}
\setcounter{algorithm}{0}

\renewcommand{\thesection}{A\arabic{section}}
\renewcommand{\thesubsection}{A\arabic{section}.\arabic{subsection}}
\renewcommand{\thefigure}{A\arabic{figure}}
\renewcommand{\thetable}{A\arabic{table}}
\renewcommand{\thealgorithm}{A\arabic{algorithm}}

\section{Additional t-SNE Visualizations}
\label{sec:appendix-tsne}

We provide t-SNE visualizations of bottom-model features for all remaining datasets. In each panel, blue points represent retained classes and red points the forgotten class. Across all datasets, BU preserves the forgotten-class cluster structure similarly to Retrain, while Target destroys global feature geometry.

\begin{figure}[h]
  \centering
  \includegraphics[width=\linewidth]{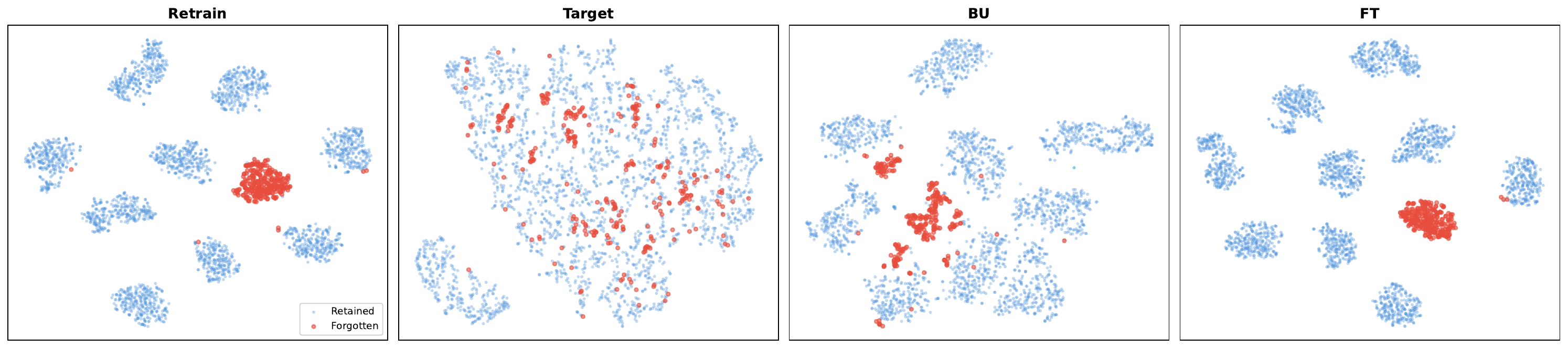}
  \caption{Bottom-model t-SNE on MNIST. BU preserves a compact forgotten-class cluster nearly identical to Retrain. Target scatters all class structure.}
  \label{fig:tsne-mnist}
\end{figure}

\begin{figure}[h]
  \centering
  \includegraphics[width=\linewidth]{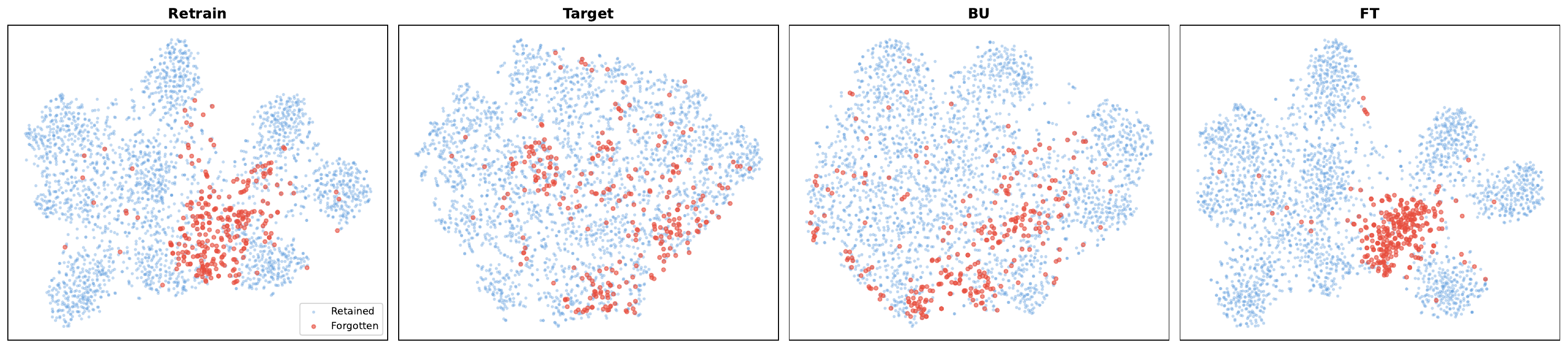}
  \caption{Bottom-model t-SNE on CIFAR-10. Despite output-level forgetting, BU retains a distinguishable forgotten-class cluster. FT preserves representational structure.}
  \label{fig:tsne-cifar10}
\end{figure}

\begin{figure}[h]
  \centering
  \includegraphics[width=\linewidth]{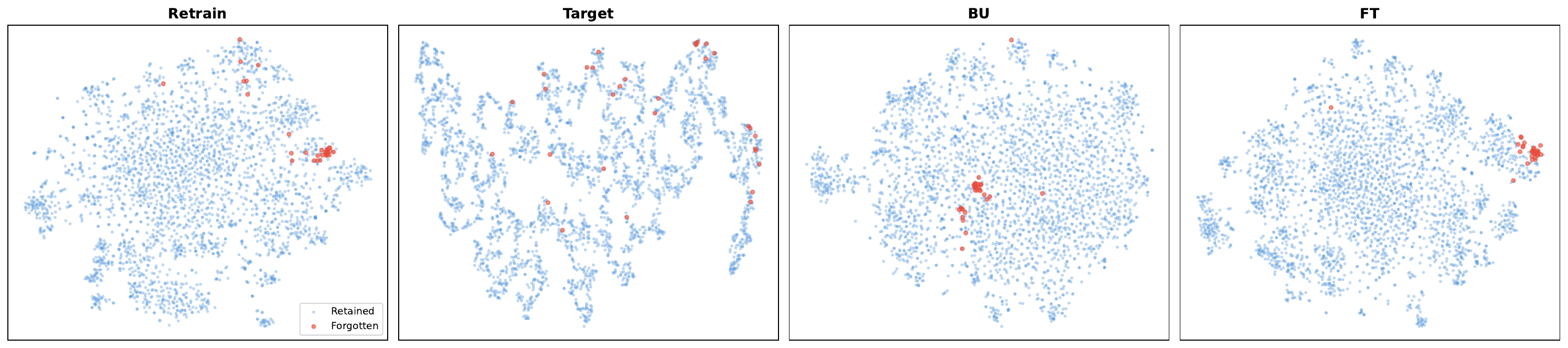}
  \caption{t-SNE visualization of bottom-model features on CIFAR-100. The forgotten class constitutes a small fraction of the total; BU preserves its spatial position relative to Retrain. Target collapses feature geometry entirely.}
  \label{fig:tsne-cifar100}
\end{figure}

\begin{figure}[h]
  \centering
  \includegraphics[width=\linewidth]{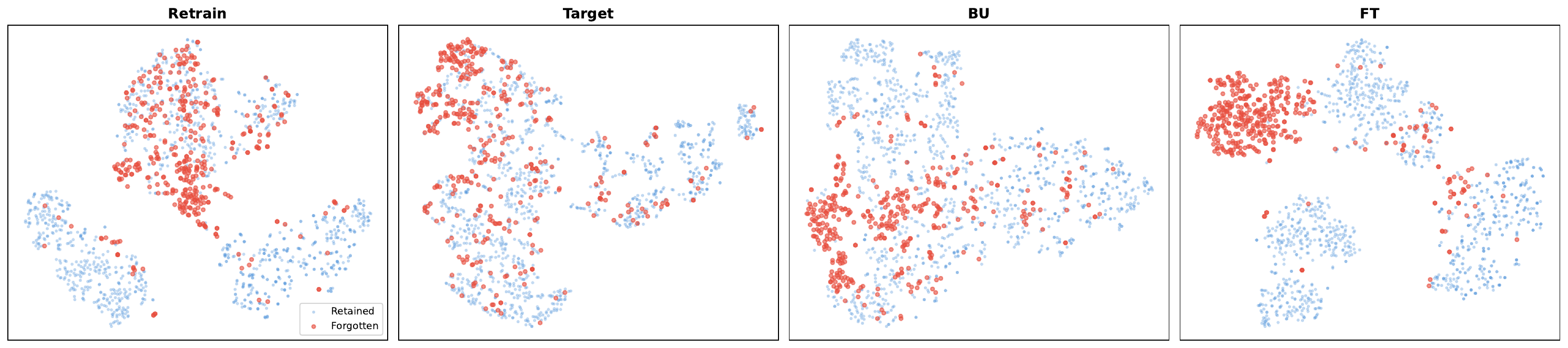}
  \caption{t-SNE visualization of bottom-model features on Brain Tumor MRI. BU maintains the forgotten-class cluster with spatial configuration close to Retrain ($\Delta_{\text{LPR}} = +8.3\%$). Target disrupts all cluster boundaries.}
  \label{fig:tsne-braintumor}
\end{figure}

\begin{figure}[h]
  \centering
  \includegraphics[width=\linewidth]{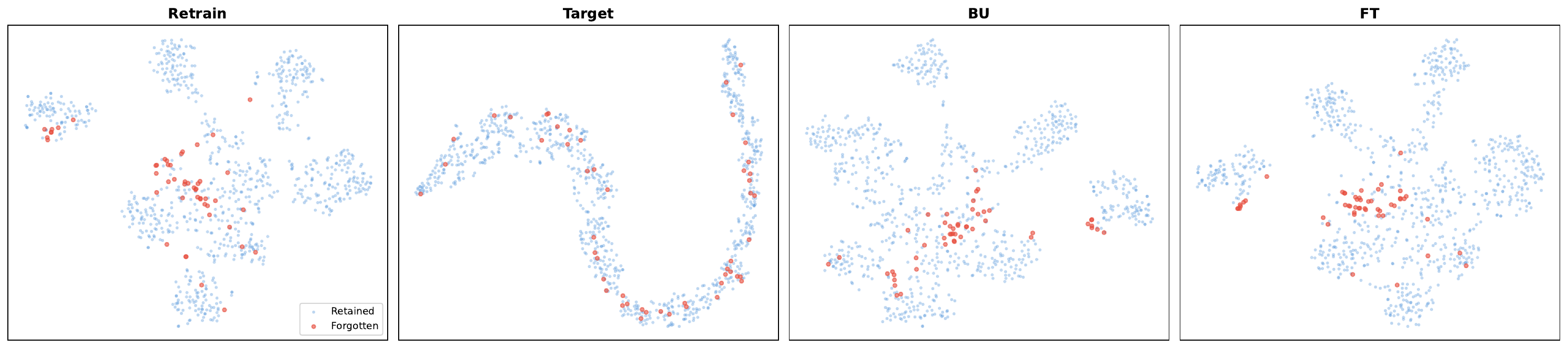}
  \caption{t-SNE visualization of bottom-model features on ModelNet. Target produces a degenerate ring structure with no class separation. BU preserves the forgotten class in a recognizable spatial arrangement.}
  \label{fig:tsne-modelnet}
\end{figure}

\begin{figure}[h]
  \centering
  \includegraphics[width=\linewidth]{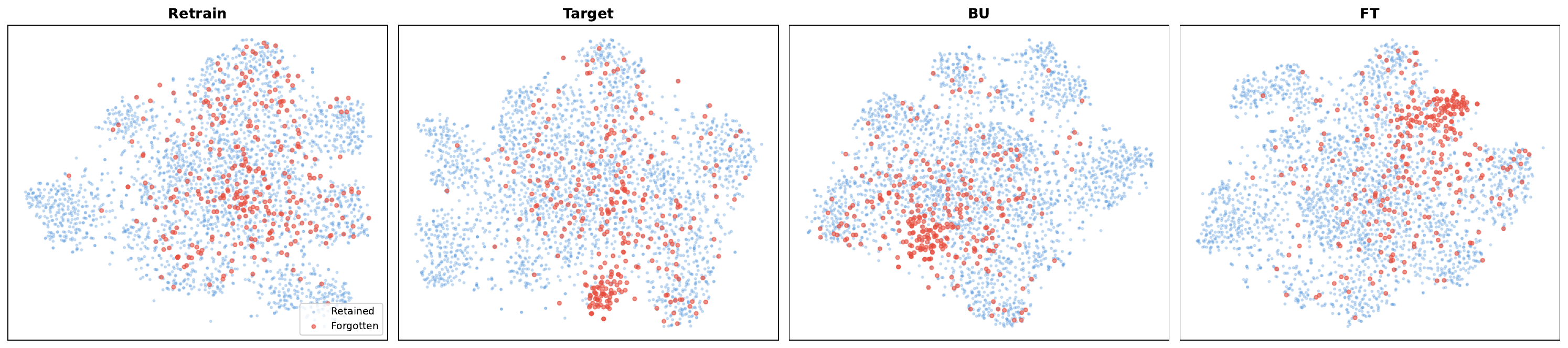}
  \caption{t-SNE visualization of bottom-model features on Yahoo Answers. As a text classification task, clusters are less compact than in image datasets. Nevertheless, BU and FT preserve the forgotten-class distribution, consistent with their positive $\Delta_{\text{LPR}}$.}
  \label{fig:tsne-yahoo}
\end{figure}

\clearpage
\section{Layer-Wise Recovery Analysis}
\label{sec:appendix-layerwise}

To validate that residual information is not confined to the penultimate layer, we evaluate LPR at three network depths: early layers (after the first residual block), intermediate layers (after the second block), and the penultimate layer (after the fourth block) of ResNet-18 on three representative datasets (see \Cref{tab:layerwise}).

\begin{table}[h]
\centering
\caption{Layer-wise LPR (\%) across network depth for three datasets (mean over 3 rounds). Early = after first residual block, Mid = after second block, Pen.\ = penultimate layer. BU and FT show higher LPR than Retrain at mid and penultimate layers, confirming that representational residuals are not confined to a single layer. Target shows LPR near or below Retrain, consistent with global structural collapse.}
\label{tab:layerwise}
\small
\setlength{\tabcolsep}{2.5pt}
\begin{tabular}{l ccc ccc ccc}
\toprule
& \multicolumn{3}{c}{\textbf{CIFAR-10}} & \multicolumn{3}{c}{\textbf{COVID-19}} & \multicolumn{3}{c}{\textbf{CIFAR-100}} \\
\cmidrule(lr){2-4} \cmidrule(lr){5-7} \cmidrule(lr){8-10}
\textbf{Method} & Early & Mid & Pen. & Early & Mid & Pen. & Early & Mid & Pen. \\
\midrule
\rowcolor{tablerowalt}
Retrain & 67.6 & 76.2 & 83.3 & 71.4 & 78.7 & 79.5 & 68.7 & 78.7 & 74.3 \\
FT      & 67.8 & 82.5 & 90.2 & 77.1 & 90.0 & 96.7 & 69.3 & 78.7 & 77.7 \\
\rowcolor{tablerowalt}
BU      & 67.1 & 80.7 & 88.1 & 76.7 & 89.9 & 95.4 & 69.3 & 78.2 & 79.4 \\
Target  & 67.6 & 78.0 & 79.4 & 66.2 & 74.5 & 68.2 & 58.2 & 58.8 & 53.8 \\
\bottomrule
\end{tabular}
\end{table}

The layer-wise results reveal that class information is distributed across the network hierarchy. On CIFAR-10, BU exhibits $\Delta_{\text{LPR}}$ of $+4.5$ at the mid layer and $+4.8$ at the penultimate layer, while FT shows even larger gaps ($+6.3$ mid, $+6.9$ penultimate). On COVID-19, the gaps are more pronounced: BU reaches $+15.9$ at the penultimate layer, with FT at $+17.2$. This confirms that residual class structure is not a shallow artifact but persists through the depth of the network. Target shows LPR below Retrain at most depths, consistent with its global collapse pattern rather than selective modification.

\clearpage
\section{Per-Class Forgetting Gap Analysis}
\label{sec:appendix-perclass}

We report per-class $\Delta_{\text{LPR}}$ for CIFAR-10 to verify that the forgetting gap is stable across different target classes, followed by a summary of results on CIFAR-100.

\begin{table}[h]
\centering
\caption{Per-class $\Delta_{\text{LPR}}$ (\%) on CIFAR-10 (mean over 3 rounds). Each column corresponds to a different forgotten class. BU maintains consistently positive gaps ($+2.3$ to $+9.1$) across all 10 classes. FT shows even larger positive gaps. Target produces values near or below zero, indicating no excess residual beyond the Retrain baseline.}
\label{tab:perclass-cifar10}
\small
\setlength{\tabcolsep}{3pt}
\begin{tabular}{l cccccccccc}
\toprule
\textbf{Method} & 0 & 1 & 2 & 3 & 4 & 5 & 6 & 7 & 8 & 9 \\
\midrule
\rowcolor{tablerowalt}
Retrain LPR & 82.6 & 86.6 & 72.7 & 71.4 & 78.7 & 75.0 & 87.0 & 83.5 & 85.8 & 87.4 \\
BU $\Delta$  & +5.5 & +5.2 & +9.1 & +4.8 & +8.1 & +5.7 & +2.3 & +6.7 & +6.2 & +3.0 \\
\rowcolor{tablerowalt}
FT $\Delta$  & +9.2 & +7.9 & +15.3 & +8.2 & +10.9 & +9.0 & +6.1 & +9.5 & +9.1 & +5.2 \\
Target $\Delta$ & $-$0.9 & $-$1.6 & $-$1.0 & $-$2.1 & $-$5.1 & $-$2.8 & $-$2.1 & $-$0.2 & +0.5 & $-$3.1 \\
\bottomrule
\end{tabular}
\end{table}

The per-class analysis confirms that the forgetting gap is a structural property of the unlearning mechanism rather than an artifact of a particular class choice. The Retrain LPR varies across classes (from $71.4\%$ for class 3 to $87.4\%$ for class 9), reflecting intrinsic differences in class separability. Despite this variation, BU consistently produces positive $\Delta_{\text{LPR}}$ for all 10 classes, with the largest gap on class 2 ($+9.1$) and the smallest on class 6 ($+2.3$). FT shows the same pattern with even larger gaps. Target, conversely, produces negative or near-zero $\Delta$ for 9 out of 10 classes, with a maximum of $+0.5$ on class 8.

On CIFAR-100, we evaluate 10 evenly spaced classes (0, 10, 20, \ldots, 90). The same pattern holds: BU produces positive $\Delta_{\text{LPR}}$ for 9 out of 10 classes (range $+0.5$ to $+10.0$), while Target is uniformly negative ($-4.8$ to $-22.7$). The Retrain LPR spans a wider range ($55.9\%$--$77.6\%$), reflecting greater heterogeneity in fine-grained class separability, but the relative method ordering remains stable.

\clearpage
\section{Nonlinear Probe Comparison}
\label{sec:appendix-nonlinear}

The main experiments use linear probes (logistic regression) to measure LPR, which provides a conservative lower bound on recoverable information. We compare linear probes with a two-layer MLP probe (hidden dimension 128, ReLU) to assess whether nonlinear methods reveal additional residual structure.

\begin{table}[h]
\centering
\caption{Linear vs.\ nonlinear (MLP) probe LPR (\%) on three datasets (mean over 3 rounds). The MLP probe achieves higher LPR than the linear probe for Retrain and FT, confirming that nonlinear residual structure exists beyond linear separability. The relative ranking of methods is largely preserved across probe types.}
\label{tab:nonlinear}
\small
\setlength{\tabcolsep}{3pt}
\begin{tabular}{l cc cc cc}
\toprule
& \multicolumn{2}{c}{\textbf{CIFAR-10}} & \multicolumn{2}{c}{\textbf{COVID-19}} & \multicolumn{2}{c}{\textbf{Brain Tumor}} \\
\cmidrule(lr){2-3} \cmidrule(lr){4-5} \cmidrule(lr){6-7}
\textbf{Method} & Linear & MLP & Linear & MLP & Linear & MLP \\
\midrule
\rowcolor{tablerowalt}
Retrain & 83.3 & 86.7 & 78.0 & 80.8 & 79.6 & 80.4 \\
FT      & 90.2 & 93.5 & 96.2 & 96.2 & 91.1 & 88.8 \\
\rowcolor{tablerowalt}
BU      & 88.1 & 87.1 & 94.6 & 92.6 & 88.9 & 86.9 \\
Target  & 79.4 & 74.3 & 67.4 & 66.7 & 79.4 & 76.0 \\
\bottomrule
\end{tabular}
\end{table}

Nonlinear probe results show that additional information is recoverable beyond linear separability for some methods. On CIFAR-10, the MLP probe increases Retrain LPR from $83.3\%$ to $86.7\%$ and FT from $90.2\%$ to $93.5\%$. For BU, the MLP probe yields comparable or slightly lower LPR than the linear probe ($88.1\%$ linear vs.\ $87.1\%$ MLP), suggesting that the residual structure captured by BU is primarily linear. Target shows consistently lower MLP LPR than linear LPR, indicating the collapsed representations do not contain meaningful nonlinear structure. These results confirm that the linear probe in the main experiments provides a conservative, reliable estimate of recoverable information.

\clearpage
\section{Computational Cost of Mirage Diagnostics}
\label{sec:appendix-cost}

We report the wall-clock time for each Mirage diagnostic tool, measured on a single NVIDIA A100 GPU with batch size 256.


\begin{table}[h]
\centering
\caption{Linear vs.\ nonlinear (MLP) probe LPR (\%) across all seven datasets (mean over 3 rounds). The method ranking is preserved across probe types. On the more separable datasets such as CIFAR-10, the MLP probe recovers higher LPR than the linear probe, confirming that the linear probe is a conservative estimate of recoverable information. On near-ceiling or lower-dimensional datasets, the two probes are comparable.}
\label{tab:nonlinear}
\setlength{\tabcolsep}{3pt}
\resizebox{\linewidth}{!}{
\begin{tabular}{l cc cc cc cc cc cc cc}
\toprule
& \multicolumn{2}{c}{\textbf{MNIST}} & \multicolumn{2}{c}{\textbf{CIFAR-10}} & \multicolumn{2}{c}{\textbf{CIFAR-100}} & \multicolumn{2}{c}{\textbf{Brain Tumor}} & \multicolumn{2}{c}{\textbf{COVID-19}} & \multicolumn{2}{c}{\textbf{ModelNet}} & \multicolumn{2}{c}{\textbf{Yahoo Answers}} \\
\cmidrule(lr){2-3} \cmidrule(lr){4-5} \cmidrule(lr){6-7} \cmidrule(lr){8-9} \cmidrule(lr){10-11} \cmidrule(lr){12-13} \cmidrule(lr){14-15}
\textbf{Method} & Lin. & MLP & Lin. & MLP & Lin. & MLP & Lin. & MLP & Lin. & MLP & Lin. & MLP & Lin. & MLP \\
\midrule
\rowcolor{tablerowalt}
Retrain & 99.4 & 98.7 & 83.3 & 86.7 & 74.3 & 71.6 & 79.6 & 80.4 & 78.0 & 80.8 & 71.6 & 65.1 & 53.6 & 54.8 \\
FT      & 99.5 & 99.6 & 90.2 & 93.5 & 77.7 & 81.1 & 91.1 & 88.8 & 96.2 & 96.2 & 77.1 & 77.1 & 60.6 & 59.9 \\
\rowcolor{tablerowalt}
BU      & 99.2 & 98.8 & 88.1 & 87.1 & 79.4 & 78.8 & 88.9 & 86.9 & 94.6 & 92.6 & 79.2 & 76.0 & 60.7 & 61.0 \\
Target  & 97.8 & 97.6 & 79.4 & 74.3 & 53.8 & 52.7 & 79.4 & 76.0 & 67.4 & 66.7 & 50.9 & 49.9 & 61.1 & 60.3 \\
\bottomrule
\end{tabular}
}
\end{table}

The Mirage audit is computationally inexpensive: $2.5$s on CIFAR-10 (50k samples), $0.3$s on COVID-19 (1.3k samples), and $0.7$s on Yahoo Answers. This represents less than $0.3\%$ of the retraining cost on CIFAR-10 and less than $1.2\%$ on Yahoo Answers. Among unlearning methods, BU and Target are the fastest ($< 2$s on COVID-19), while FT requires multiple passes through the retained data. The total cost of applying Mirage to audit any unlearned model is dominated by the unlearning procedure itself; the diagnostic overhead is negligible.

\clearpage

\section{Extended Discussion}
\label{sec:appendix_discussion}

\paragraph{Representation-Level Erasure as a Structural Constraint.}
Our results show that the gap between output-level forgetting and representation-level erasure is structural rather than incidental. In deep models, class identity is encoded in the geometry of feature space through stable cluster structure and inter-class margins. Adjusting classifier weights or suppressing logits modifies decision boundaries but does not necessarily reshape this geometry. Because linear recoverability depends on feature separability rather than output logits, forgotten-class information can remain decodable even when predictions collapse. 
This motivates a retraining-relative criterion. A retrained model captures the intrinsic feature geometry induced by the retained data. A positive forgetting gap ($\Delta_{\text{LPR}} > 0$) therefore indicates residual structure beyond what retraining would preserve. In this sense, representation-level forgetting requires geometric alignment with the retrained reference rather than suppressing output behavior.

\paragraph{An Empirical Unlearning Trilemma.}
Across all datasets and methods, we observe a consistent tension among three desiderata: 
(1)~\textbf{utility} (high retained accuracy), 
(2)~\textbf{output forgetting} (low forgotten-label accuracy), and 
(3)~\textbf{representation forgetting} (small forgetting gap $\Delta_{\text{LPR}}$). 
None of the evaluated methods achieves all three simultaneously. Methods that preserve utility often retain recoverable class structure, while methods that strongly suppress outputs tend to degrade retained performance through global feature distortion. Although we do not claim a formal impossibility result, the consistency of this pattern across datasets suggests a fundamental structural tension between preserving feature geometry and eliminating class separability.

\paragraph{Class--Sample Asymmetry.}
Class-level and sample-level unlearning exhibit different representational behaviors. Class identity corresponds to global geometric structure in feature space and therefore remains linearly recoverable from penultimate-layer embeddings. In contrast, individual sample membership does not appear as a stable linear signal in the representation space. Our experiments show that linear probes reliably recover class information but fail to identify specific forgotten samples once labels are removed.
This observation suggests that auditing mechanisms should match the granularity of the forgetting objective. Class-level erasure requires geometry-aware diagnostics such as LPR, CKA alignment, and separability analysis. Sample-level auditing, however, may require alternative tools such as membership inference, influence estimation, or data attribution methods.

\end{document}